\documentclass[10pt,twocolumn,letterpaper]{article}

\usepackage{cvpr}              

\usepackage{graphicx}
\usepackage{amsmath}
\usepackage{amssymb}
\usepackage{booktabs}
\usepackage[export]{adjustbox}

\usepackage[pagebackref,breaklinks,colorlinks]{hyperref}

\usepackage{xcolor}

\newcommand\blfootnote[1]{%
  \begingroup
  \renewcommand\thefootnote{}\footnote{#1}%
  \addtocounter{footnote}{-1}%
  \endgroup
}

\usepackage[capitalize]{cleveref}
\crefname{section}{Sec.}{Secs.}
\Crefname{section}{Section}{Sections}
\Crefname{table}{Table}{Tables}
\crefname{table}{Tab.}{Tabs.}

\usepackage{multirow}

\def\ours{LayoutDiffuse}
\def\oursshort{LayoutDiff.}

\def\cvprPaperID{10488} 
\def\confName{CVPR}
\def\confYear{2023}

\begin{document}

\title{LayoutDiffuse: Adapting Foundational Diffusion Models for Layout-to-Image Generation}

\author{Jiaxin Cheng\textsuperscript{*1}, Xiao Liang\textsuperscript{*2}, Xingjian Shi\textsuperscript{3}, Tong He\textsuperscript{3}, Tianjun Xiao\textsuperscript{3}, Mu Li\textsuperscript{3}\\
\textsuperscript{1}USC Information Sciences Institute, \textsuperscript{2}Shanghai Jiao Tong University, \textsuperscript{3}Amazon Web Services\\
{\tt\small chengjia@isi.edu, liang\_xiao@sjtu.edu.cn, \{xjshi,htong,tianjux,mli\}@amazon.com}}

\twocolumn[{%
\renewcommand\twocolumn[1][]{#1}%
\maketitle
\vspace{-25pt}
\begin{center}
    \centering
    \includegraphics[width=\textwidth,trim=0.0cm 0.5cm 0.0cm 0.15cm,clip]{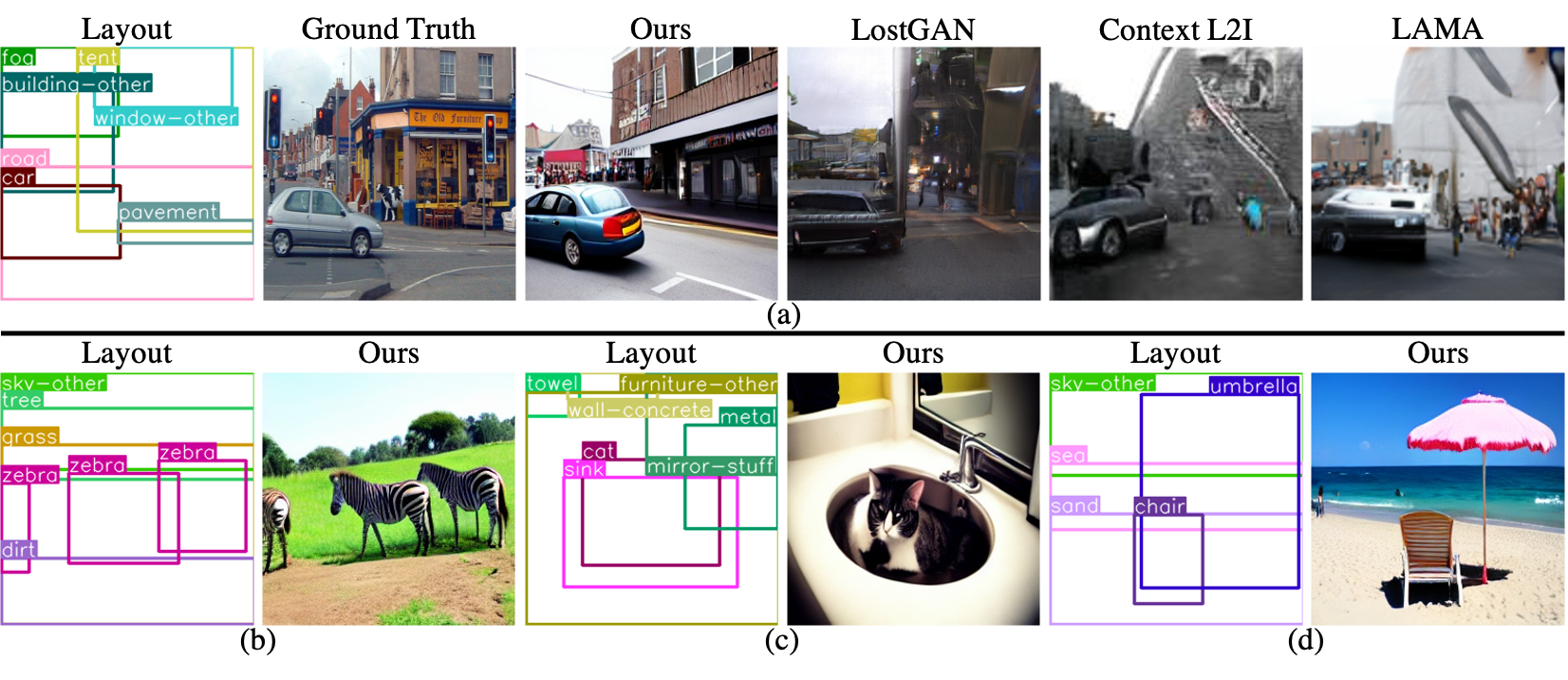}
    \captionof{figure}{\label{fig.teaser} \ours\ is able to generate high quality images that aligns well with complex layouts. (a) Bounding box layout-to-image comparison between \ours~and LostGAN~\cite{lostgan}, Context L2I~\cite{context_l2i} and LAMA~\cite{lama}. (b), (c) and (d) Box-to-image generation in different scenes. (b) Animals in the wild. (c) Indoor. (d) Outdoor.}\label{fig.teaser}
\end{center}%
}]

\begin{abstract}
   Layout-to-image generation refers to the task of synthesizing photo-realistic images based on semantic layouts. In this paper, we propose \ours\  that adapts a foundational diffusion model pretrained on large-scale image or text-image datasets for layout-to-image generation. By adopting a novel neural adaptor based on layout attention and task-aware prompts, our method trains efficiently, generates images with both high perceptual quality and layout alignment, and needs less data.  Experiments on three datasets show that our method significantly outperforms other 10 generative models based on GANs, VQ-VAE, and diffusion models.
   
\end{abstract}


\section{Introduction}


Layout-to-image generation aims to synthesize photo-realistic images based on the semantic layout. In this paper, we define layout as either a segmentation mask~\cite{spade} or a collection of labeled object bounding boxes~\cite{lostgan}. Compared with other conditional signals like the tokenized class label~\cite{ho2022classifier} or text description of the image~\cite{ramesh2021zero}, the layout is easy to get and captures important coarse-level semantic structure like the type and location of the objects. This offers users more control of the final looking of the image and can bring new ways of computer-assisted content creation.

\blfootnote{\textsuperscript{*}This work was completed during his internship at Amazon.} 
Despite its importance, layout-to-image generation is challenging because the generated image needs to be both perceptually plausible and consistent with the layout. Previous works often adopt Generative Adversarial Networks (GANs)~\cite{spade,lee2020maskgan,lama,pSp,lostgan,context_l2i} as the building block for layout-to-image generation models. However, as illustrated in \cref{fig.teaser}, these models fail to generate reasonable images when the layout is complex. On the other hand, we are witnessing a breakthrough in image synthesis brought by the advance of Diffusion Models (DMs)~\cite{ddpm,ldm}. DMs generate the image by iteratively refining the noisy signal with a likelihood function modeled via a U-Net. Since DM is based on maximizing the likelihood~\cite{ldm}, it does not have the mode collapse problem of GANs~\cite{gan} and has been shown to better capture semantic information~\cite{ddim} and generate images with better sample quality than GANs. DMs have been applied to conditional generation tasks like text-to-image generation~\cite{dalle2, imagen}, image inpainting~\cite{palette}, and super-resolution~\cite{saharia2022image}. However, few works studied how to extend DMs for layout-to-image generation. Rombach et al.~\cite{ldm} proposed the pioneering Latent Diffusion Model (LDM), which applies DM in the latent space of a Vector-Quantized Variational Auto-Encoder (VQ-VAE)~\cite{vqvae}. This work tokenized the layout to a sequence of coordinates and object classes and applied the text-to-image DM for layout-to-image generation. However, the distribution of the pseudo-text converted from layout is far from the distribution of the real-world text that LDM was trained on. Thus such approach cannot leverage a pretrained text-to-image LDM and requires the costly process of re-training the full model.


In this paper, we propose to adapt the pretrained image-only or text-to-image DMs~\cite{ldm}, of which we call \emph{foundational diffusion models}, for layout-to-image generation. Our algorithm, called \emph{\ours}, implants the layout signal in DM by adding our newly designed neural adapter~\cite{houlsby2019parameter, jia2022visual}. The adapter contains two components: layout attention and task-adaptive prompts. Layout attention concentrates the self-attention to be within each instance to emphasize the context. Correlations among instances are modeled by globally shared class embeddings. Task-adaptive prompts are tunable vectors with the purpose of informing the foundational DM to switch to the layout-to-image generation mode. To avoid hampering the performance of the pretrained DM and accelerating the convergence, we inject adapters as residual blocks so that they will not impact model performance at the initial stage of fine-tuning.

Our contributions are summarized as follows: 1) we propose layout attention and task-adaptive prompts for adapting foundational DMs for layout-to-image generation; 2) by adapting a foundational DM, we are able to reduce the training time of layout-to-image models from days to hours; 3) our neural-adapter-based method is data-efficient and can achieve good performance with a small number of training samples; 4) \ours{} significantly outperforms prior methods on three datasets: CelebA-Mask~\cite{CelebAMask-HQ}, COCO Stuff~\cite{caesar2018coco}, and Visual Genome~\cite{vg}.



\section{Related works}

\noindent\textbf{Layout-to-image generation}: layout-to-image generation can be seen as a reverse process of instance segmentation or object detection. Early works~\cite{lostgan,ocgan,lama,context_l2i,spade,pSp} often adopt GANs~\cite{gan} for such tasks while recently the diffusion models~\cite{palette,ldm} have started to show promising results and draw the community's attention. SPADE~\cite{spade} and MaskGAN~\cite{lee2020maskgan} adopt encoder-decoder architecture for transferring mask to image, and \cite{scene_graph,zhao2019image} use similar architecture to generate images from bounding boxes. These methods encode the layout as an image that is downsampled and upsampled jointly with the data. 
StyleGAN~\cite{karras2019style} architecture is used in pSp~\cite{pSp} for mask-to-image.  LAMA~\cite{lama}, LostGANs~\cite{lostgan}, Context L2I~\cite{context_l2i} encode layout as a style feature and feed into an adaptive normalization layer.
Taming~\cite{taming} and TwFA~\cite{twfa} encode layout information as inputs to transformer~\cite{allyouneed} and use auto-regressive (AR) transformer to predict the latent visual codes from the pretrained VQ-VAE~\cite{vqvae}. More recent diffusion model-based works~\cite{palette,ldm}) use segmentation mask as condition by simply concatenating the mask to the input image alone channel dimension. LDM~\cite{ldm} uses a text-to-image model for box-to-image transfer by encoding layout using a BERT model. Unlike these papers, we encode layouts by adjusting the self-attention mask to focus on the instances and adding prompt tokens.

\noindent\textbf{Diffusion models}~\cite{ddpm,improved_ddpm,score_diffusion} are generative models that synthesize images from random noise by iterative image denoising.  DDGAN~\cite{ddgan}, DiffusionVAE~\cite{diffusevae,diffuse_vae2} study the combination of diffusion model and other generative methods. DDIM~\cite{ddim} and PLMS~\cite{plms} mitigate the lengthy sampling procedure by reducing number of sampling steps down to dozens from thousands. LDM~\cite{ldm} leverages VQ-VAE~\cite{vqvae} to encode images to latent codes with smaller resolution, saving efforts to train super-resolution models for generating high-resolution images like~\cite{imagen,glide,dalle2}. DreamBooth~\cite{dreambooth} and TextualInversion~\cite{textual_inversion} study fine-tuning diffusion model to produce variant images of the same object. Make-a-video~\cite{make_a_video} fine-tunes text-to-image diffusion models for text-to-video generation by using spatial attention across frames. These fine-tuning strategies, however, require the same type of conditional signal for fine-tuning as in the pretraining, \ie, a text conditioned model still need text after fine-tuning. In contrast, \ours\ can easily adapt conditional models to be unconditioned, enabling our method for a wider range of applications.


\section{Methodology}

\begin{figure}
    \centering
    \includegraphics[width=\linewidth]{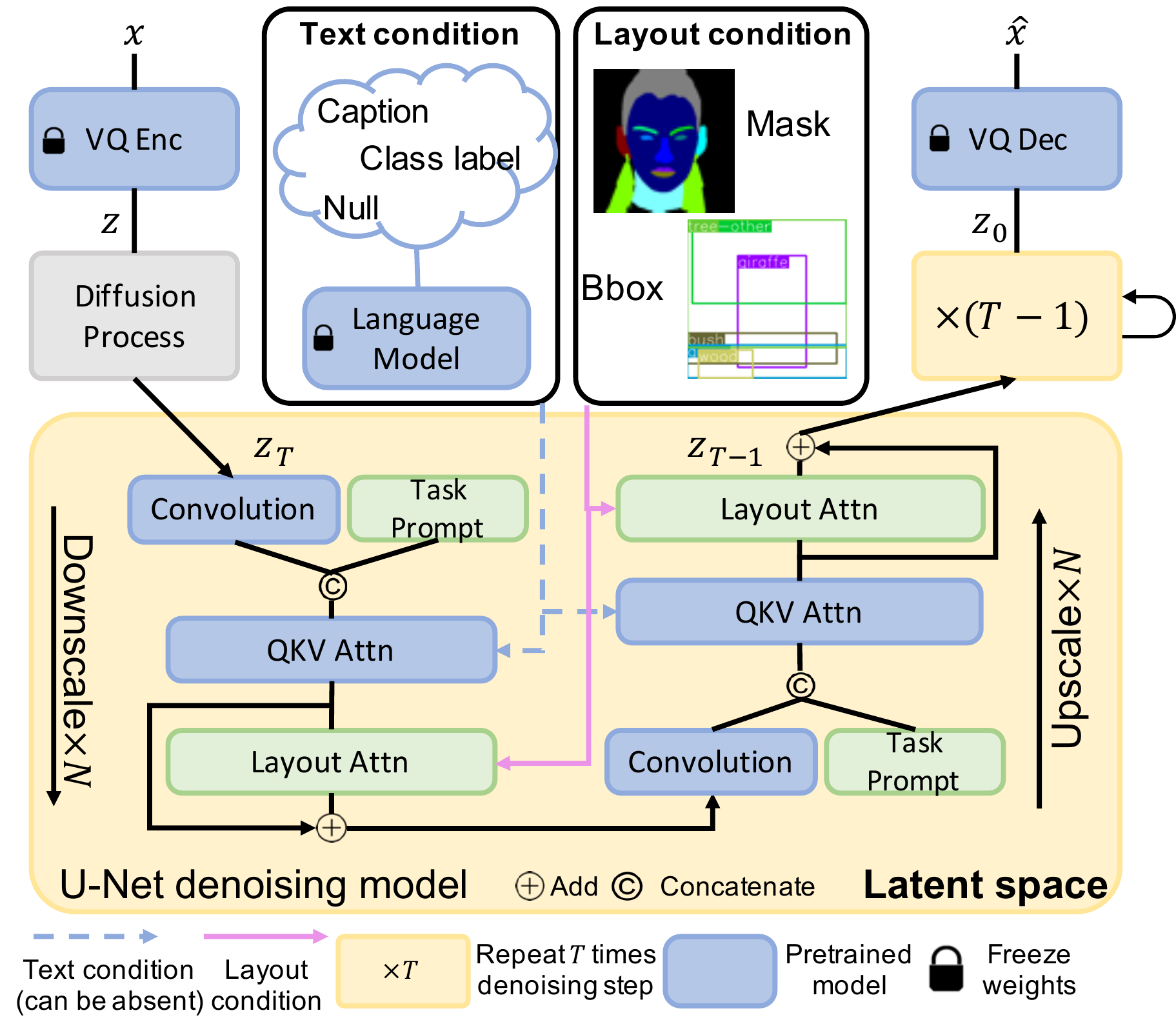}
    \caption{The overall architecture of \ours. We plugin task-adaptive prompts and layout attention layers to the pretrained diffusion model. The task-adaptive prompts (\cref{sec.task_prompt}) are added to QKV attention layers, and the layout attention layers (\cref{sec.instance_prompt_attn}) are designed as residual blocks. The pretrained diffusion model can be either conditioned on text or unconditional.}
    \label{fig.archi}
\end{figure}

\subsection{Overall}

\ours\ adapts a foundational DM, either text-conditioned or unconditioned, for layout-to-image generation. We leverage the latent diffusion model (LDM)~\cite{ldm} as our backbone, which has demonstrated the capability to generate high-quality images with less computational cost. We propose two components, layout attention (\cref{sec.instance_prompt_attn}) and task-adaptive prompts (\cref{sec.task_prompt}), to adapt the pretrained DM to be layout conditioned. The overall architecture is illustrated in~\cref{fig.archi}. Here, layout attention adjusts the attention mask to be aware of the instances in the layout and is added as a residual block on top of the QKV attention in the backbone. Task-adaptive prompts provide additional context to the QKV attention layers~\cite{allyouneed}.

In our experiments, unless explicitly clarified, we train the whole model end-to-end except the VQ-VAE and text encoder in LDM. Thus, even without task-adaptive prompts and the instance-aware component in layout attention, the model can learn to condition on the encoded layout. However, we find that adding these two components helps model achieve better performance within the same number of tuning steps. In addition, we find that the layout attention layer helps the model generate images that are both more faithful to the layout and have better perceptual quality, especially in the bounding box layout-to-image generation setting.



\subsection{Diffusion Backbones}
\noindent\textbf{Diffusion Models} (DM)\cite{ddpm,improved_ddpm} are generative models that learns to predict the distribution of data $p(x)$ from normally distributed random noise vectors. Its superiority over classic generative methods (\eg VAE~\cite{vae}, GANs~\cite{gan}) has been validated on conditional~\cite{glide,imagen,dalle2,ldm,palette} and unconditional generation~\cite{ddpm,improved_ddpm,beat_gan} tasks. It samples an image by a progressive $T$-step noise reduction from the initial random noise. The training process of DM can be interpreted as the inverse process of a Markov chain of length $T$~\cite{ddpm}. The model optimizes the variational lower bound of $p(x)$~\cite{ddpm,beat_gan,improved_ddpm}, which includes $T$ networks $\varepsilon_\theta(x_t, t)$ predicting the original noise $\varepsilon$ from the data $x_t$ corrupted after $t$ diffusion steps.




\begin{equation}
    L = \mathbb{E}_{x,\varepsilon\sim\mathcal{N}(0,1),t}[\|\varepsilon - \varepsilon_\theta(x_t, t)\|^2].
\end{equation}

\noindent\textbf{Latent Diffusion Model} (LDM)~\cite{ldm}, instead of generating images via DM in the RGB pixel space, generate the visual codes extracted via a pretrained Vector Quantized Variational AutoEncoder (VQ-VAE)~\cite{vqvae}. LDM enables the generation of high-resolution images with less computational resources and higher quality. Specifically, the image $x$ is first encoded to latent space $z=\mathrm{VQEnc}(x)$. The diffusion model is trained to predict the distribution $p(z)$ instead of $p(x)$. Therefore, we replace $x_t$ with $z_t$ in \cref{eq.ldm_objective}. The predicted latent codes are finally decoded back to RGB space by $\hat{x}=\mathrm{VQDec}(z)$:
\begin{equation}\label{eq.ldm_objective}
    L = \mathbb{E}_{\mathrm{VQEnc}(x),\varepsilon\sim\mathcal{N}(0,1),t}[\|\varepsilon - \varepsilon_\theta(z_t, t)\|^2].
\end{equation}


\subsection{Initializing from Foundational Models}
Training DMs from scratch can take weeks~\cite{beat_gan} even with thousands of GPUs. Our strategy of fine-tuning pretrained DMs greatly reduces the training workload to dozens of hours with a few GPUs. We bring layout conditioning in the image generation by adding layout attention layers after QKV attention in the backbone. We design layout attention layers to be residual blocks with an output linear layer initialized to zero. Such design ensures the layer is an identity mapping at the start of fine-tuning. Therefore, all convolutional layers and QKV attention layers (see \cref{fig.archi}) can be initialized with pretrained weights, and the initial model behaves the same as the pretrained foundational DM.


\begin{figure}
    \centering
    \includegraphics[width=0.85\linewidth]{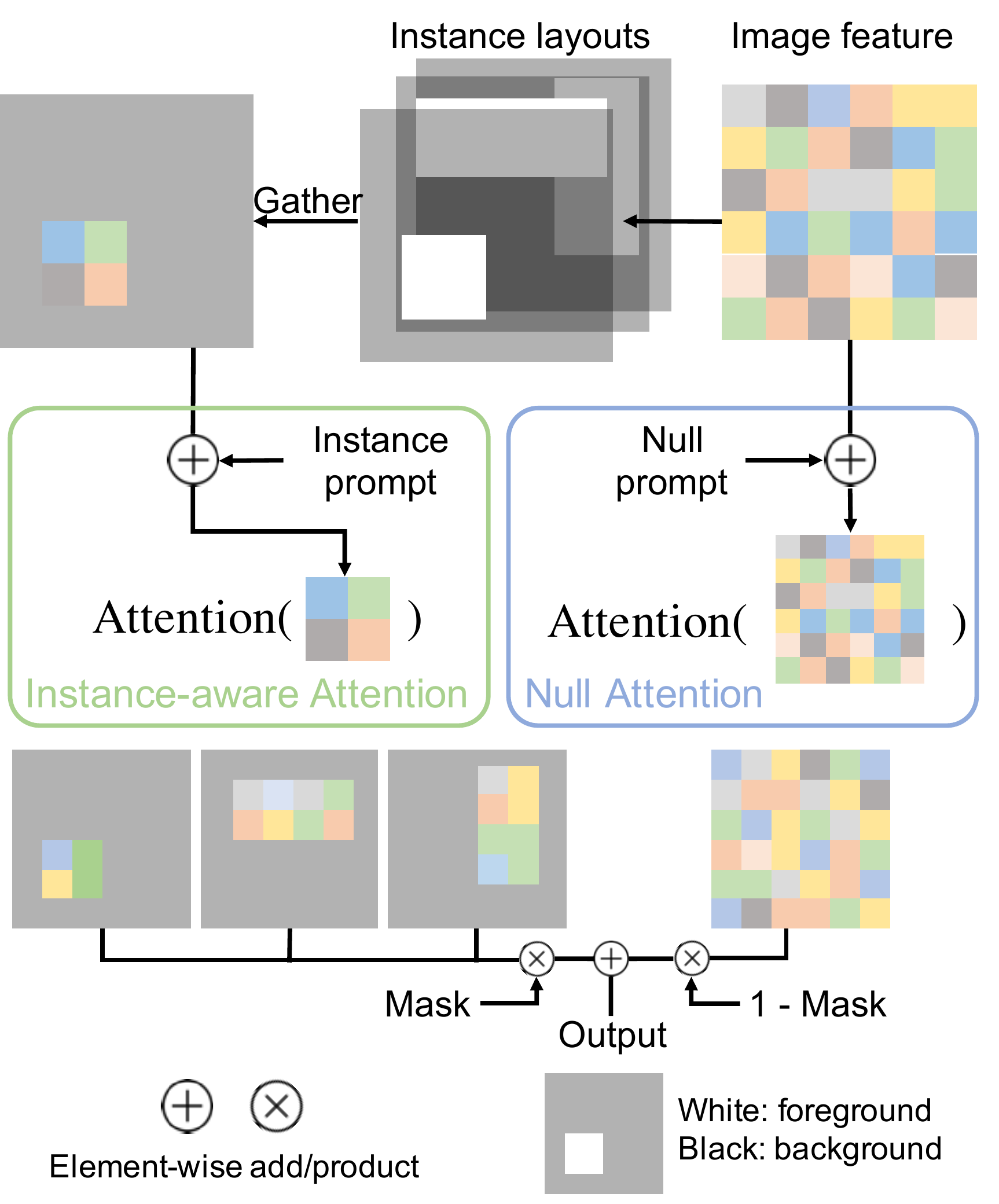}
    \caption{The illustration of the layout attention layer. The null attention is also used in the classifier free guidance as a negative condition. ``Mask'' indicates foreground object locations.}
    \label{fig.layout_attention}
\end{figure}

\subsection{Layout Attention}
\label{sec.instance_prompt_attn}

The layout attention layer incorporates layout information in the DM backbone. It has two key modifications over the normal self-attention: \textit{instance prompt} (see \cref{fig.layout_attention}). To inform model the location and category of the instances, we mark each instance by adding a learnable token to the image features in the region containing the instance. Specifically, we set instance prompt as a learnable class embedding $e_k$ for each category $k$ in the data, and the instance prompt will be added if region $r$ contains $k$ during instance-aware attention. Instance-aware attention is a regional self-attention layer where each token only attends to other tokens that belong to the same instance. Let $a^{(i)}$ be the feature region of $i$-th instance with class $k$ in the image, the instance-aware attention obtains the query, key, and value as
\begin{equation}
    Q_r^{(i)} = \varphi_Q(r^{(i)}), K_r^{(i)} = \varphi_K(r^{(i)}), V_r ^{(i)}= \varphi_V(r^{(i)}),
\end{equation}
\noindent where $r^{(i)} = a^{(i)} + e_k$ is the regional image feature after adding class embedding, $\varphi_{Q},\varphi_{K},\varphi_{V}$ are linear projection layers shared across instances. For the region $r^{\varnothing}$ that contains no instance (\ie, the background area), we learn a null instance prompt $e_\varnothing$ and apply attention on tokens belong to the background $\mathrm{Attention}\left(\varphi_{Q}(r^{(\varnothing)}), \varphi_K(r^{\varnothing}), \varphi_V(r^{\varnothing})\right)$ where $r^{\varnothing} = a^{\varnothing} + e_\varnothing$. We merge the results of instance-aware attention and null attention according to the foreground and background mask. For overlapping objects, we average the features after instance-aware attention as we empirically found doing so performs better than averaging class embeddings before attention. The learnable class embeddings are shared across all layout attention layers. 

The null embedding is also used for classifier-free guidance (CFG)\cite{ho2022classifier}. CFG was proposed for improving the sample quality of conditional DMs. During CFG, the DM predicts denoised target of both positive condition (\eg, text prompt) and negative condition (\eg, an empty string for text prompt). The final prediction is then extrapolated from negative towards positive with a coefficient $s$ greater than 1. More precisely, let $\varepsilon$ and $\varepsilon_{\varnothing}$ be the predicted noise of positive and negative conditions, the prediction after CFG can be expressed as~\cref{eq.cfg}.  In our case, we use an empty layout as the negative condition and the instance-aware attention turns into a global self-attention with null embeddings.
\begin{align}\label{eq.cfg}
    \varepsilon'=(1-s)\times \varepsilon_{\varnothing} + s \times (\varepsilon - \varepsilon_{\varnothing})
\end{align}

\subsection{Task-adaptive Prompts}\label{sec.task_prompt}

Task-adaptive prompts are cues for the model to recognize if the generation task has been changed to layout-to-image from the pretraining task (\ie, unconditional image generation or text-to-image generation). Instead of adding prompts to input/intermediate features like Visual Prompt Tuning~\cite{vpt}, we add task-adaptive prompts into the standard QKV attention~\cite{allyouneed}, $\mathrm{Attention}(Q, K, V) = \mathrm{softmax}(QK^T/\sqrt{d})V$ by concatenating the prompts to keys and values. Specifically, let $a \in \mathbb{R}^{l\times d}$ be the input image feature to the attention layer, $g\in \mathbb{R}^{m\times d}$ be the task-adaptive prompts that contain $m$ learnable embeddings of dimension $d$, we have 
\begin{equation}
    Q = \varphi_Q(a), K = \varphi_K([a, g]), V = \varphi_V([a, g]),
\end{equation}
\noindent where $l=h\times w$ and $d$ are the spatial and feature dimension of flattened image feature, $[\cdot]$ concatenates features along the spatial dimension so that $[a, g]\in\mathbb{R}^{(l+m)\times d}$. The task-adaptive prompts are added to all QKV attention layers in the model and are initialized with Gaussian distribution. Since the feature dimension is unchanged, the QKV attention can still work with pretrained weights. We empirically chose $m=64$ as the number of prompts as we found adding more prompts only gives marginal improvement.

In the case where the cross attention is used for conditional generation~\cite{imagen,ldm,glide,dalle2}, we concatenate the task-adaptive prompts to conditional embedding $c \in \mathbb{R}^{r\times d}$ instead, where $r$ is $c$'s spatial dimension (\cref{eq.cross_attention}). Moreover, via task-prompt, we are able to convert a conditional DM to an unconditional one, \eg, converting a text-to-image model to text-free layout-to-image model. To do so, we remove the text condition model and set condition $c=\varnothing$
\begin{equation}\label{eq.cross_attention}
    Q = \varphi_Q(a), K = \varphi_K([c, a, g]), V = \varphi_V([c, a, g])
\end{equation}


\begin{figure*}[!ht]
    \centering
    \includegraphics[width=\linewidth,trim=0.0cm 4.5cm 0.0cm 0cm,clip]{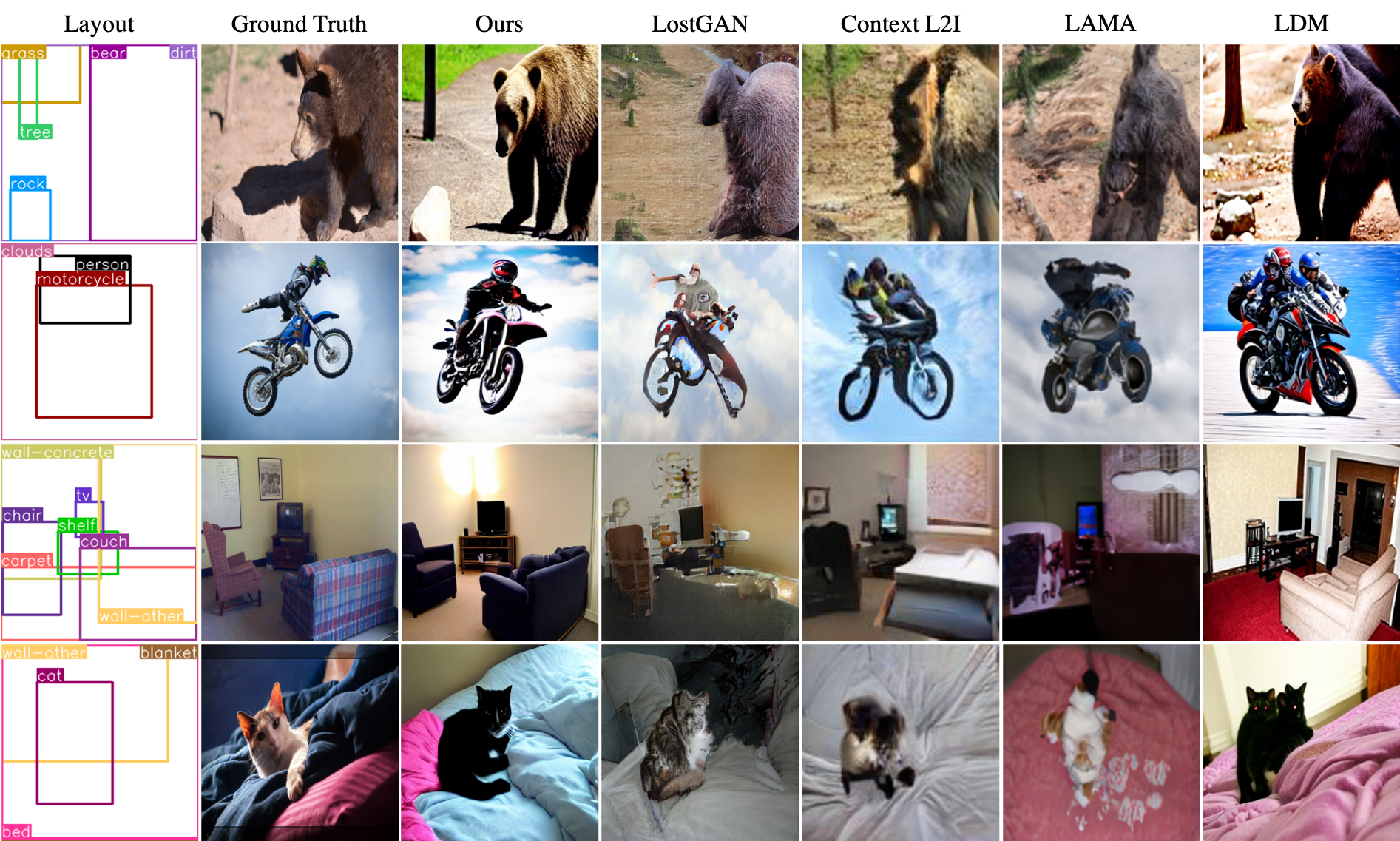}
    \caption{Qualitative comparison of bounding box layout-to-image. Our method can generate high-quality images that fits complex layouts and the generated images have much better quality than GAN-based methods and align to the layout better than LDM.}
    \label{fig.bbox_qualitative}
    \vspace{-1em}
\end{figure*}

\begin{figure*}[!ht]
    \centering
    \includegraphics[width=\linewidth]{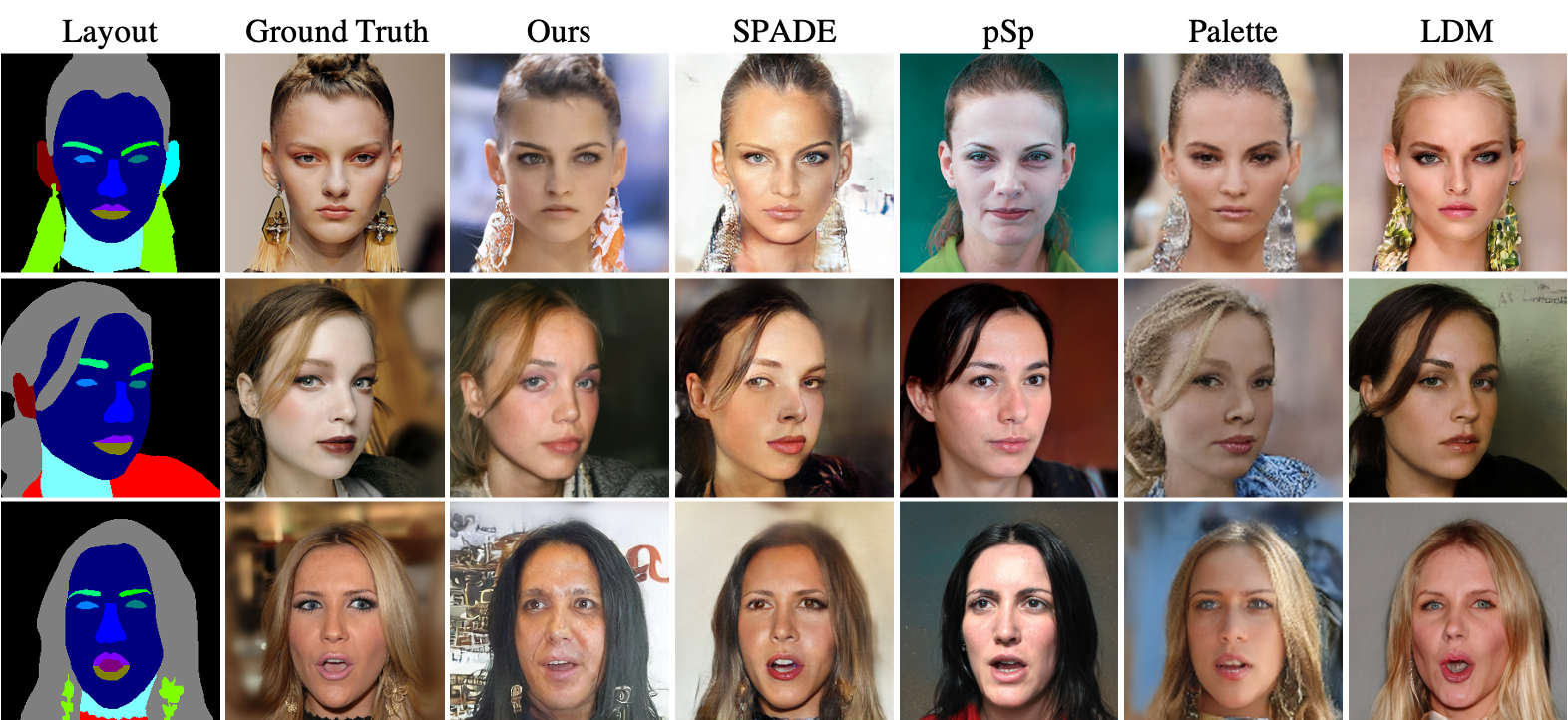}
    \caption{Qualitative comparison of mask layout-to-image. Our model is fine-tuned with only 30 epochs. Palette and LDM are trained with 500 and 200 epochs, respectively.}
    \label{fig.seg_qualitative}
    \vspace{-1em}
\end{figure*}

\section{Experiments}

\subsection{Dataset}

\noindent\textbf{Bounding box layout-to-image}: we assess bounding box layout-to-image performance with COCO Stuff~\cite{caesar2018coco} and Visual Genome (VG)~\cite{vg}. We follow previous works~\cite{lama} to filter the number of objects in an image from 3 to 8 in COCO and 3 to 30 in VG during test. We also eliminate the objects that are smaller than 2\% of image area, resulting in 112,680 training and 3,097 testing samples in COCO from 171 classes, and 62,565 training and 5,062 testing samples in VG from 178 categories. 

\noindent\textbf{Mask layout-to-image}: we use CelebA-Mask~\cite{CelebAMask-HQ} dataset for evaluating mask layout-to-image performance. The dataset has 30,000 facial images and corresponding semantic masks, covering 19 facial attribute classes. We benchmark the performance on 2,993 test split images. 

\subsection{Metrics}
\noindent\textbf{Bounding box layout-to-image} Quantitatively, we evaluate the perceptual quality of the generated image via Frechet Inception Distance (FID)~\cite{fid}, Inception Score (IS)~\cite{inception_score}, Classification Accuracy Score (CAS)~\cite{lama}, and evaluate whether the generated image aligns well with layout via the YOLO score~\cite{lama} and SceneFID~\cite{ocgan}. FID measures the generation quality by comparing the statistical distance between generated images and real images. IS is a metric for both image quality and diversity, which has been noticed to correlate well with human evaluation. CAS score measures the quality of image by training an auxiliary classifier on generated objects and evaluate its performance on real objects. YOLO score compares the ground-truth and inferred layout of the generated image, of which a higher value indicates the image better aligns with layout. SceneFID computes the FID on crops of all objects and is tailored to the layout-to-image generation task. We follow~\cite{lama} to train a ResNet-101 on generated object crops and test on validation crops. 
In addition, we conduct human evaluation on the layout-to-image generation algorithms on 50 participants using COCO-based results. We ask participants to choose one preferred image among baseline methods~\cite{context_l2i,lama,lostgan,ldm} listed in~\cref{tab.benchmark} and \ours{}. Each participant is given 15 questions. Images in each question are conditioned on the same layout and resized to $256 \times 256$ resolution with randomly shuffled order. For each layout and the corresponding set of images, users will be asked to either choose the image with the best overall quality, or choose the image with the best layout fidelity. Also, user can choose ``None of above'' if there is no satisfactory result.

\noindent\textbf{Mask layout-to-image} is evaluated on FID and mean Intersection of Union (mIoU). The mIoU is computed between generated image and conditioning ground truth mask using a pretrained face parsing model~\cite{dml_csr}. Higher mIoU indicates that the generated image is more recognizable and aligns better to the conditioning mask. 

\subsection{Implementation Details}\label{sec.implementation_details}
Our box-to-image model is initialized from the text-to-image LDM pretrained on LAION-400M~\cite{laion} by~\cite{ldm}, which has a denoising U-Net/VQ-VAE/language model of 903M/83.7M/581M parameters. The mask-to-image model is initialized from an unconditional facial model pretrained on CelebA-Mask~\cite{CelebAMask-HQ} by~\cite{ldm}, having denoising U-Net/VQ-VAE of 280M/55.3M parameters. We fine-tune models with Adam~\cite{adam} optimizer for 30, 60 and 120 epochs at learning rate $3\times 10^{-5}$ on CelebA-Mask, COCO and VG, respectively. We train models on NVIDIA A10G GPUs, taking 50 hours for COCO and VG on 24 GPUs, and 10 hours for CelebA-Mask on 4 GPUs. The effective batch size is 128 after applying gradient accumulation. We sample images using PLMS~\cite{plms} sampler for 100 steps with classifier-free guidance~\cite{ho2022classifier} scale at $5.0$. We inject layout attention layers after QKV attention layers when the image resolution is 32 and 16 as we empirically find applying layout attention to smaller resolution gives marginal improvement. 

For adapting the text-to-image model, we use the caption from COCO~\cite{caesar2018coco} as the text prompt. Our experiments show that the performance is similar when the caption is not used (\cref{sec.ablation_archi}). VG~\cite{vg} does not have image caption, we instead use comma-connected class labels (\eg, ``car,tree,building'') as image caption.

\subsection{Baselines}
We compare to GAN-based, VQ-VAE+AR-based and diffusion-based methods for layout-to-image generation. ~\Cref{tab.baseline_methods} summarizes the baseline methods that are used in the experiments. For all GAN-based and VQ-VAE+AR methods, we use the released pretrained model from official implementation. When the implementation is not available, we report numbers in the original paper. For diffusion-based methods, we follow ~\cite{ldm} to convert layout to conditional signals. The bounding box information is encoded by a language model after tokenizing the object and location, and the segmentation information is encoded by concatenating the segmentation mask and image along the channel dimension. To ensure fair comparison, the baseline LDM~\cite{ldm} is initialized from the same pretrained weights as \ours. We fine-tune LDM for 60, 120 epochs on COCO and VG, and 200 epochs on CelebA-Mask.

\begin{table*}[]
    
    \begin{minipage}[t]{.59\textwidth}
        \centering
        \scriptsize
        \begin{tabular}{l@{\hspace{3pt}}c@{\hspace{3pt}}c@{\hspace{3pt}}c@{\hspace{3pt}}c@{\hspace{3pt}}c@{\hspace{3pt}}c@{\hspace{3pt}}c}
        \toprule
        Backbone & \multicolumn{4}{c}{GAN} & \multicolumn{2}{c}{VQVAE+AR}  & \multicolumn{1}{c}{Diffusion} \\
        \cmidrule(lr){1-1} \cmidrule(lr){2-5} \cmidrule(lr){6-7} \cmidrule(lr){8-8} 
        Method & LostGAN\cite{lostgan} & OCGAN\cite{ocgan} & LAMA\cite{lama} & Context L2I\cite{context_l2i} & Taming\cite{taming} & TwFA\cite{twfa} & LDM\cite{ldm} \\
        Year & 2019 & 2021 & 2021 & 2021 & 2022 & 2022 & 2022 \\
        \bottomrule
        \end{tabular}
    \end{minipage}%
    \hspace{30pt}
    \begin{minipage}[t]{.39\textwidth}
        \raggedright
        \scriptsize
        \begin{tabular}{l@{\hspace{3pt}}c@{\hspace{3pt}}c@{\hspace{3pt}}c@{\hspace{3pt}}c@{\hspace{3pt}}c}
        \toprule
        Backbone & \multicolumn{2}{c}{GAN}   & \multicolumn{2}{c}{Diffusion} \\
        \cmidrule(lr){1-1} \cmidrule(lr){2-3} \cmidrule(lr){4-5} 
        Method & SPADE\cite{spade} & pSp\cite{pSp} & Palette\cite{palette} & LDM\cite{ldm}  \\
        Year & 2019 & 2021 & 2022 & 2022 \\
        \bottomrule
        \end{tabular}
    \end{minipage}
    \label{tab.baseline_methods}
    \caption{Baseline methods for bounding box layout-to-image (left) and segmentation layout-to-image (right)}
\end{table*}

\subsection{Results}
\cref{tab.benchmark} summarizes the quantitative results of baseline methods and \ours{} on bounding box layout-to-image generation. We notice that \ours\ achieves SoTA on all three metrics and is the most preferred algorithm in human evaluation. \ours\ achieves a significant improvement on CAS and IS, suggesting that the objects generated by our method are more similar to real objects (higher CAS) and are more diversified and distinguishable (higher IS).

In \cref{fig.bbox_qualitative}, we qualitatively compare the generated images of different methods. Samples in the same row have the same layout conditioning. It is easy to see that images generated by \ours\ are more clear and align better with the ground truth layout. 

\ours\ also achieves SoTA on mask-to-image generation and the results are presented in \cref{tab.result_celebmask}. Overall, the diffusion-based methods achieve better performance. Palette~\cite{palette}, though being a diffusion-based method, does not perform as well as \ours\ and LDM, which can largely be due to the fact that it is trained directly in RGB pixel space, which is more difficult. Compare to LDM, \ours\ achieves better performance with much less training epochs. \cref{fig.seg_qualitative} demonstrates that \ours\ can produce better quality images that fit well with conditioning masks with much less training efforts.

\begin{table*}[!t]
    
    \centering
    \begin{tabular}{lcccccccc}
    \toprule
    \multirow{2}*{\textbf{Methods}} & \multicolumn{2}{c}{\textbf{FID$\downarrow$}} & \multicolumn{2}{c}{\textbf{CAS$\uparrow$}} & \multicolumn{2}{c}{\textbf{Inception Score$\uparrow$}} & \multicolumn{2}{c}{\textbf{User Preference$\uparrow$}}\\
     \cmidrule(lr){2-3} \cmidrule(lr){4-5} \cmidrule(lr){6-7}  \cmidrule(lr){8-9} 
    & COCO & VG & COCO & VG & COCO & VG & Quality & Align Fidelity \\
    \cmidrule(lr){1-1} \cmidrule(lr){2-2} \cmidrule(lr){3-3} \cmidrule(lr){4-4} \cmidrule(lr){5-5} \cmidrule(lr){6-6} \cmidrule(lr){7-7} \cmidrule(lr){8-8} \cmidrule(lr){9-9} 
    LostGAN~\cite{lostgan} & 42.55 & 47.62 & 30.33 & 28.81 & 18.01\small{$\pm$0.50} & 14.10\small{$\pm$0.38} & 5.22\% & 2.50\% \\
    OCGAN*~\cite{ocgan} & 41.65 & 40.85 & - & - & - & - & - & - \\
    Context L2I~\cite{context_l2i} & 29.56 & 119.74 & 32.63 & 20.09 & 18.57\small{$\pm$0.54} & 6.71\small{$\pm$0.30} & 1.10\% & 2.50\% \\
    LAMA~\cite{lama} & 31.12 &  31.63  & 30.52 & 31.75 & 14.32\small{$\pm$0.58} & 10.79\small{$\pm$0.66} & 1.98\% & 3.21\% \\
    Taming*~\cite{taming} & 33.68 & 19.14 & - & - & - & - & - & - \\
    TwFA*~\cite{twfa} & 22.15 & 17.74 & - & - & 24.25\small{$\pm$1.04} & 25.13\small{$\pm$0.66} & - & - \\
    LDM~\cite{ldm} & 24.60 & 25.27 & 43.48 & 34.78 & 26.11\small{$\pm$0.88} & 20.59\small{$\pm$0.41} & 20.88\% & 24.38\% \\
    \cmidrule(lr){1-1} \cmidrule(lr){2-3} \cmidrule(lr){4-5} \cmidrule(lr){6-7}
    \cmidrule(lr){8-9}
    \ours~(Ours) & \textbf{20.27} & \textbf{15.96} & \textbf{50.02} & \textbf{40.41} & \textbf{32.07\small{$\pm$0.80}} & \textbf{26.53\small{$\pm$0.41}} & \textbf{62.36\%} & \textbf{61.88\%} \\
    \bottomrule
    \end{tabular}
    \label{tab.benchmark}
    \caption{Bounding box layout-to-image results on COCO and VG. `$\uparrow$' stands for higher the better, `$\downarrow$' stands for lower the better. All generated images are evaluated under $256\times256$ resolution. Our user study on COCO collects feedback from 50 users. Numbers from methods marked with `*' are copied from the original paper.}
    \vspace{-1em}
\end{table*}

\begin{table}[h]
    
    \centering
    \begin{tabular}{lcc}
    \toprule
    \textbf{Methods} & \textbf{FID$\downarrow$} &\textbf{mIoU(\%)$\uparrow$} \\
    \cmidrule(lr){1-1} \cmidrule(lr){2-2} \cmidrule(lr){3-3}
    SPADE~\cite{spade} & 29.86 & 60.37  \\
    pSp~\cite{pSp} & 62.78 & 53.95  \\
    Palette (@500 epoch)~\cite{palette} & 33.82 & 60.96 \\
    LDM (@200 epoch)~\cite{ldm} & 19.44 & 64.17 \\
    \cmidrule(lr){1-1} \cmidrule(lr){2-3}
    \oursshort~(@30 epoch) & \textbf{8.11} & \textbf{67.05} \\
    \bottomrule
    \end{tabular}
    \caption{Mask-to-image on CelebA-Mask. \ours\ can achieve better performance with only 30 epochs fine-tuning. }
    \label{tab.result_celebmask}
\end{table}

\subsection{Object Recognizability}
We use YOLO score~\cite{lama} and SceneFID~\cite{lama} to evaluate the recognizability of the object in the generated images. YOLO score uses a pretrained YOLOv4~\cite{yolo} model to detect the objects in the generated images. This metric reflects both the generation quality as well as the alignment fidelity to the reference layout. SceneFID is the FID score computed on the cropped objects, which measures the distribution difference between real and generated objects.  Better quality images should generate easier-to-recognize objects and therefore have higher YOLO score and lower SceneFID. We report Average Precision (AP), AP50, AP75, Average Recall (AR) and SceneFID in~\cref{tab.map_mar}. Our method outperforms the best non-diffusion baseline by a large margin. When comparing to LDM, \ours\ still obtains more than 4\%/6\% improvement regarding AP/AR, illustrating the efficiency and effectiveness of our method.

\begin{table}[!h]
    
    \centering
    \scriptsize
    \begin{tabular}{lccccc}
    
    \toprule
    \textbf{Methods} & \textbf{AP$\uparrow$} &\textbf{AP50$\uparrow$} & \textbf{AP75$\uparrow$} & \textbf{AR$\uparrow$} & \textbf{SceneFID$\downarrow$} \\
    \cmidrule(lr){1-1} \cmidrule(lr){2-2} \cmidrule(lr){3-3} \cmidrule(lr){4-4} \cmidrule(lr){5-5} \cmidrule(lr){6-6}
    LAMA & 20.61 & 28.54 & 22.69 & 26.48 & 18.64 \\
    LostGAN & 17.10 & 28.50 & 18.82 & 22.98 & 22.00 \\
    Context L2I & 10.03 & 14.90 & 11.14 & 12.71 & 14.40 \\
    Taming & - & - & - & - & 13.36 \\
    TwFA & - & 28.20 & 20.12 & - & 11.99 \\
    LDM & 32.13 & 54.23 & 34.34 & 39.86 & 17.36 \\
    \cmidrule(lr){1-1} \cmidrule(lr){2-6}
    \ours & \textbf{36.58} & \textbf{59.59} & \textbf{38.06} & \textbf{46.09} & \textbf{11.92} \\
    \bottomrule
    \end{tabular}
    \caption{Object recognizability result on COCO. The objects generated by \ours\ align with layout better (higher YOLO score) and are more realistic (lower SceneFID).}
    \label{tab.map_mar}
\end{table}

\subsection{Ablation Studies}

\subsubsection{Model Architecture}\label{sec.ablation_archi}

\begin{figure}[!h]
    \centering
    \includegraphics[width=\linewidth]{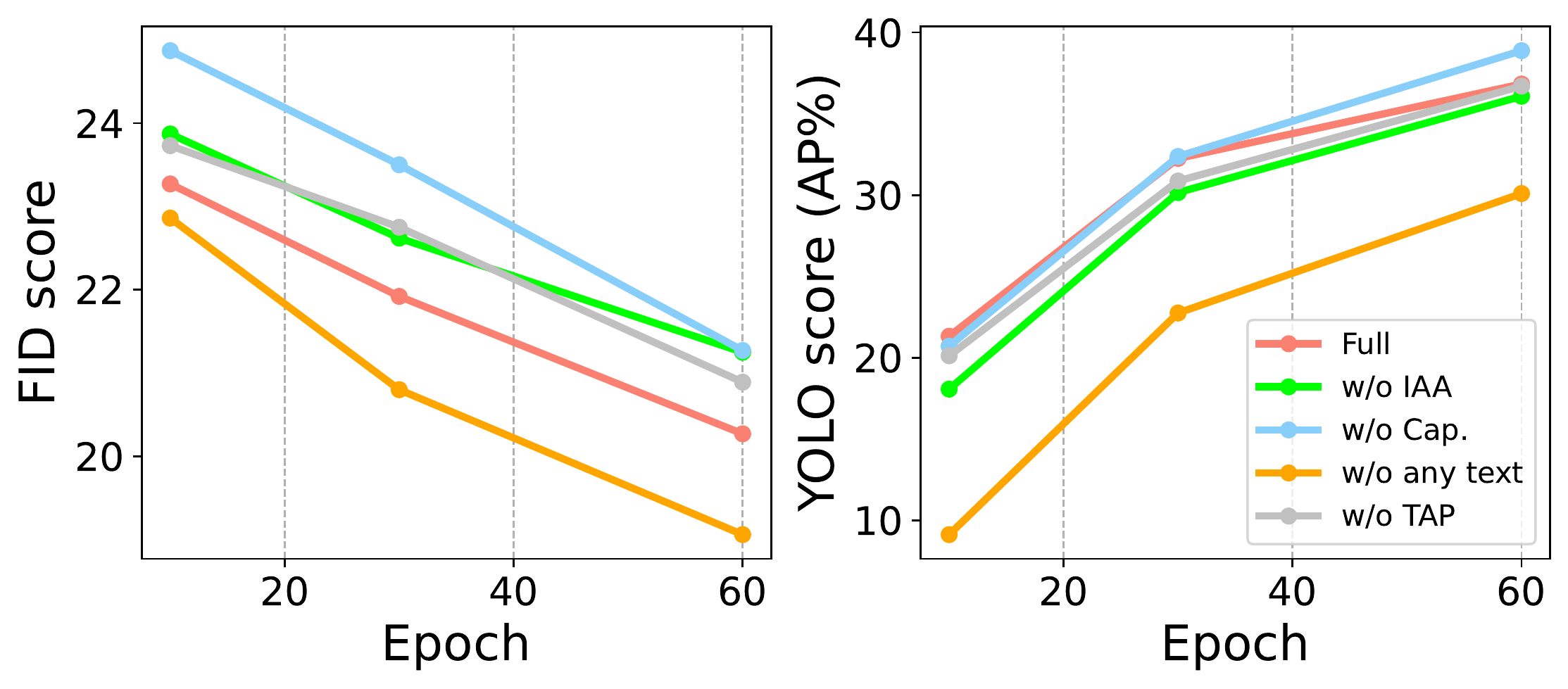}
    \caption{FID and YOLO score (Average Precision) of fine-tuned model on COCO. IAA: instance-aware attention, Cap.: image caption, TAP: task-adaptive prompts. See \ref{fig:qualitative_fig_6} for qualitative comparison at different epochs.}
    \label{fig.model_ablation}
    \vspace{-1em}
\end{figure}

We study the effectiveness of task-adaptive prompt, instance-aware attention and text condition by removing these components and compare with the full model. When instance-aware attention is removed, we apply global attention on the whole image feature after adding instance prompts. For the text condition, we conduct two experiments, i) removing captions and using comma-connected class labels of the given bounding boxes as mentioned in~\cref{sec.implementation_details} and ii) removing the whole text conditioning model. We denote two captionless cases as \textit{w/o Cap.} and \textit{w/o any text}. In the second case, we reinitialize the cross-attention layer where text tokens were used and replace the text tokens with the concatenation of image feature and task-adaptive prompts as mentioned in~\cref{sec.task_prompt}. 

We train all models on COCO under the same setting as mentioned in \cref{sec.implementation_details}. We compare the FID and YOLO Average Precision scores, and the results are presented in~\cref{fig.model_ablation}. We can see that when training with instance-aware attention and task-adaptive prompts, the model always obtains better performance on both FID and YOLO scores. When evaluated on the YOLO score, the \textit{full} model has similar final performance as the captionless models. However, adding instance-aware attention and task-adaptive prompt improves fine-tuning efficiency, especially at early epochs.

Captionless training, however, shows a trade-off between image quality and fidelity to the layout with two different trendings. When training without any text prompt, the FID score is significantly lower than other models, at the sacrifice of a worse YOLO score. We conjecture that this is because the foundational model is a text-to-image model. Removing the text model will cause the loss of image content (\ie, harder to recognize object) and, as a result, the training optimizes more on overall image quality (\ie, model learns global distribution better). On the other hand, using image label only gives better YOLO score with worse FID. We speculate that explicitly providing image labels pushes the model to focus on generating objects (higher YOLO), thus holding back the optimization speed for improving image quality (worse FID). 

\subsubsection{Fine-tuning Efficiency}\label{sec.efficiency}

The most important motivation for fine-tuning rather than training from scratch is that fine-tuning is time and data efficiency. We leverage CalebA-Mask~\cite{CelebAMask-HQ} dataset to illustrate the above two aspects of our fine-tuning strategy.

\begin{figure}[!h]
    \centering
    \includegraphics[width=\linewidth]{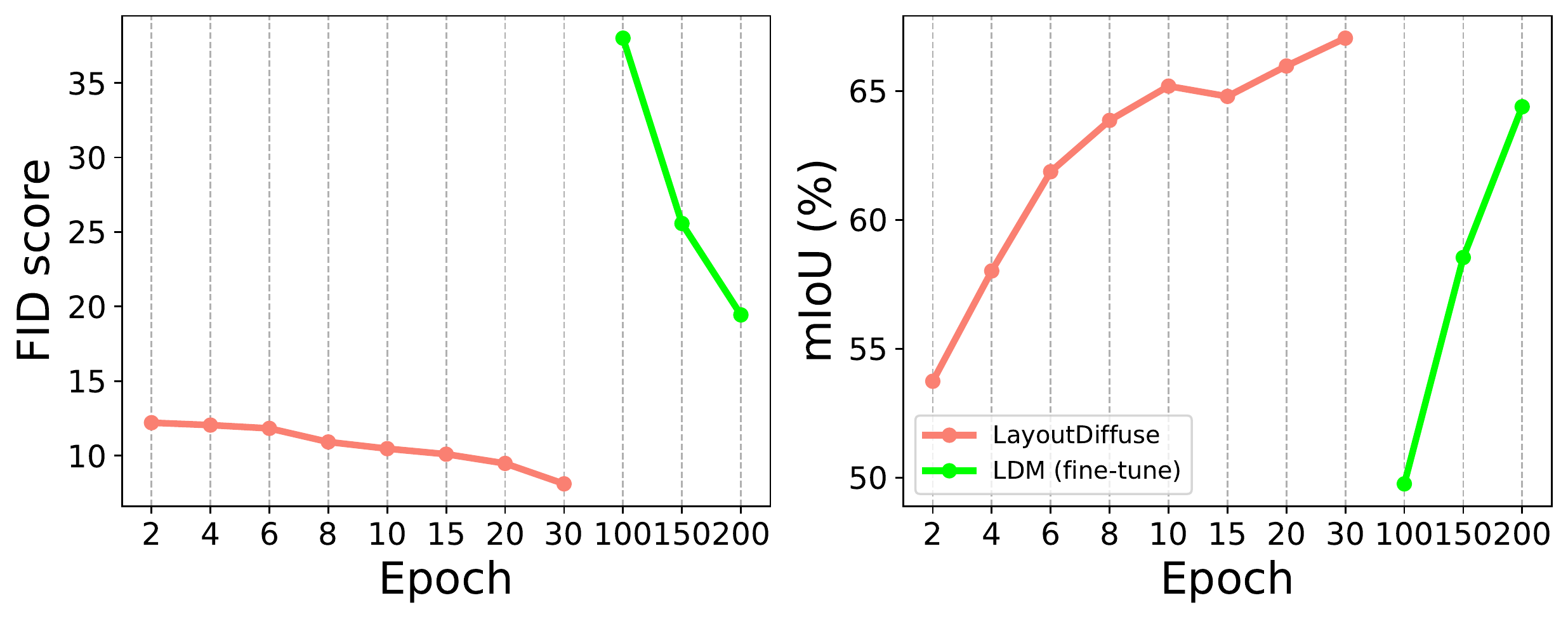}
    \caption{FID and mIoU score of fine-tuned model from unconditional model to mask layout-to-image model on CelebA-Mask. Our model is fine-tuned up to 30 epochs. The baseline LDM model is fine-tuned for 200 epochs from the same initial weights. Please note that the $x$-axis has been rescaled for better visualization. See \ref{fig:qualitative_figure_7} for qualitative comparison at different epochs.}
    \label{fig.time_efficiency}
\end{figure}

We study \textbf{time efficiency} by training the model up to 30 epochs. As a comparison, we initialize LDM from the same pretrained weights and only replace its input convolutional kernel from 3 to 6 channels after concatenating the segmentation mask to the diffused image. Only the channels of segmentation mask are randomly initialized. \cref{fig.time_efficiency} compares the FID and mIoU score of \ours\ and fine-tuned LDM. By inserting layout attention as a residual network, \ours\ is able to generate images from a good initial point, and the only objective the model learns is adding the condition to the image. Padding the conditioning mask to the input along the image channel, however, can hardly leverage the foundational model. The fine-tuned LDM takes 200 epochs to achieve similar performance on mIoU and cannot produce the same quality image regarding the FID score. We think this is because the input distribution is completely changed after adding the conditioning mask, and the model needs to learn everything from the beginning. 

\begin{figure}[!h]
    \centering
    \includegraphics[width=\linewidth]{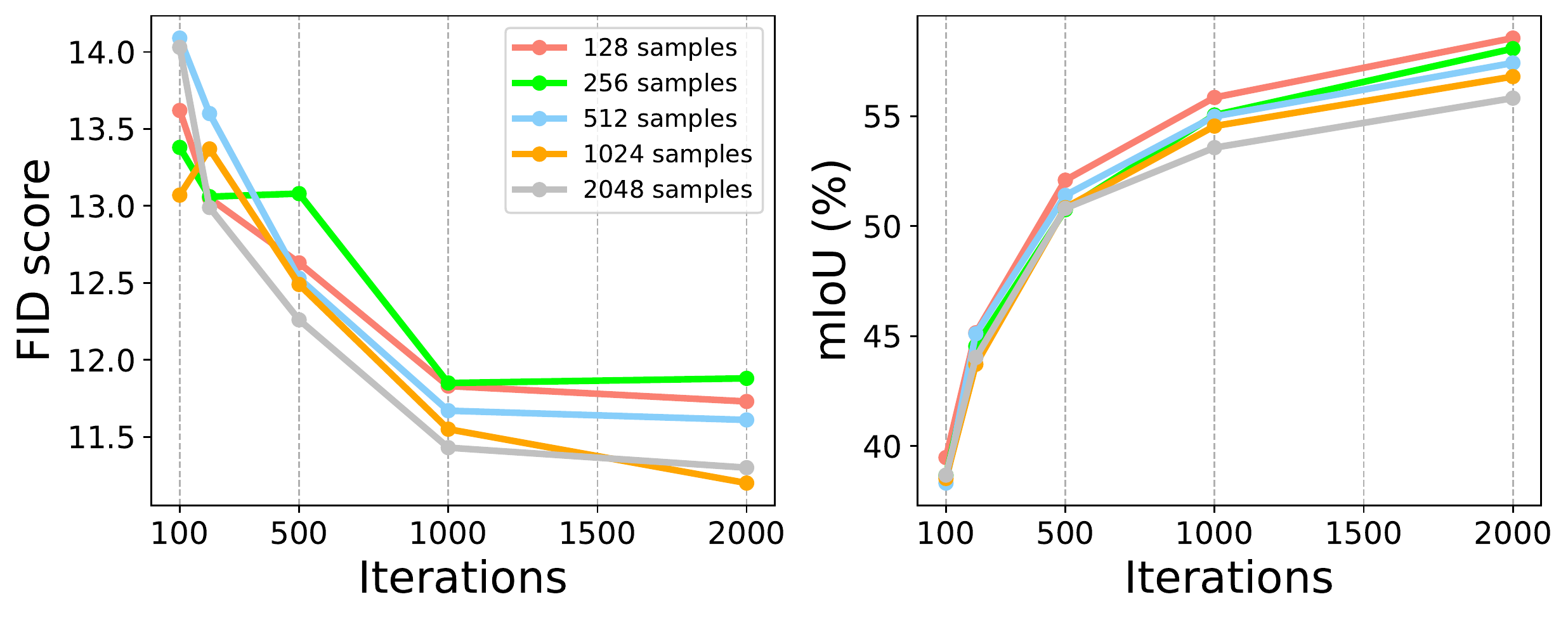}
    \caption{FID and mIoU scores of the fine-tuned model from unconditional model to mask layout-to-image model on CelebA-Mask. Iteration is counted at batch size 128. 2,000 iterations approximate to 10 epochs training if using full dataset.}
    \label{fig.data_efficiency}
\end{figure}

To illustrate \textbf{data efficiency}, we randomly pick 128, 256, 512, 1024 and 2048 images for fine-tuning. In this experiment, we freeze all the pretrained weights to avoid model collapsing. \cref{fig.data_efficiency} shows the performance of fine-tuning after 100, 200, 500, 1,000 and 2,000 iterations. We noticed that the final FID and mIoU score is worse than the full data training. We attribute this to the pretrained weights being frozen. However, \ours\ is still able to generate high-quality images with low FID and high layout fidelity, with mIoU comparable to non-diffusion baselines. Similar to the above results when studying the effect of using image caption, we observe a trade-off between image quality and layout fidelity. Both FID score and mIoU decreas as the number of training images increases. We suspect using more training samples causes worse mIoU because the layout conditioning is a new task to learn. Therefore, more training data requires more time for the model to adapt and, as a result, under the same number of iterations, fewer training samples obtain higher mIoU scores. 

\section{Conclusion}

We present \ours\ to fine-tune a foundational diffusion model, either trained on text-image pairs or only on images, to be layout conditioned. We propose task prompt and instance prompt for speeding up model adaptation, which are compatible to both mask and bounding box layout-to-image generation. Our design of layout attention layer as a residual block allows us to adapt foundational model with minimum fine-tuning efforts. Our experimental results show that \ours\ i) generates SoTA high quality images with ii) more recognizable objects and iii) is time and data efficient.

{\small
\bibliographystyle{ieee_fullname}
\bibliography{egbib}
}
\clearpage
\appendix
\twocolumn[{%
\renewcommand\twocolumn[1][]{#1}%
\begin{center}
\textbf{\Large LayoutDiffuse: Adapting Foundational Diffusion Models for Layout-to-Image Generation}

\vspace{10pt}
\textbf{\large Supplementary Material}
\vspace{10pt}
\end{center}%
}]

\section{Importance of Null Condition in Classifier Free Guidance}
\begin{figure}[!h]
    \centering
    \includegraphics[width=\linewidth]{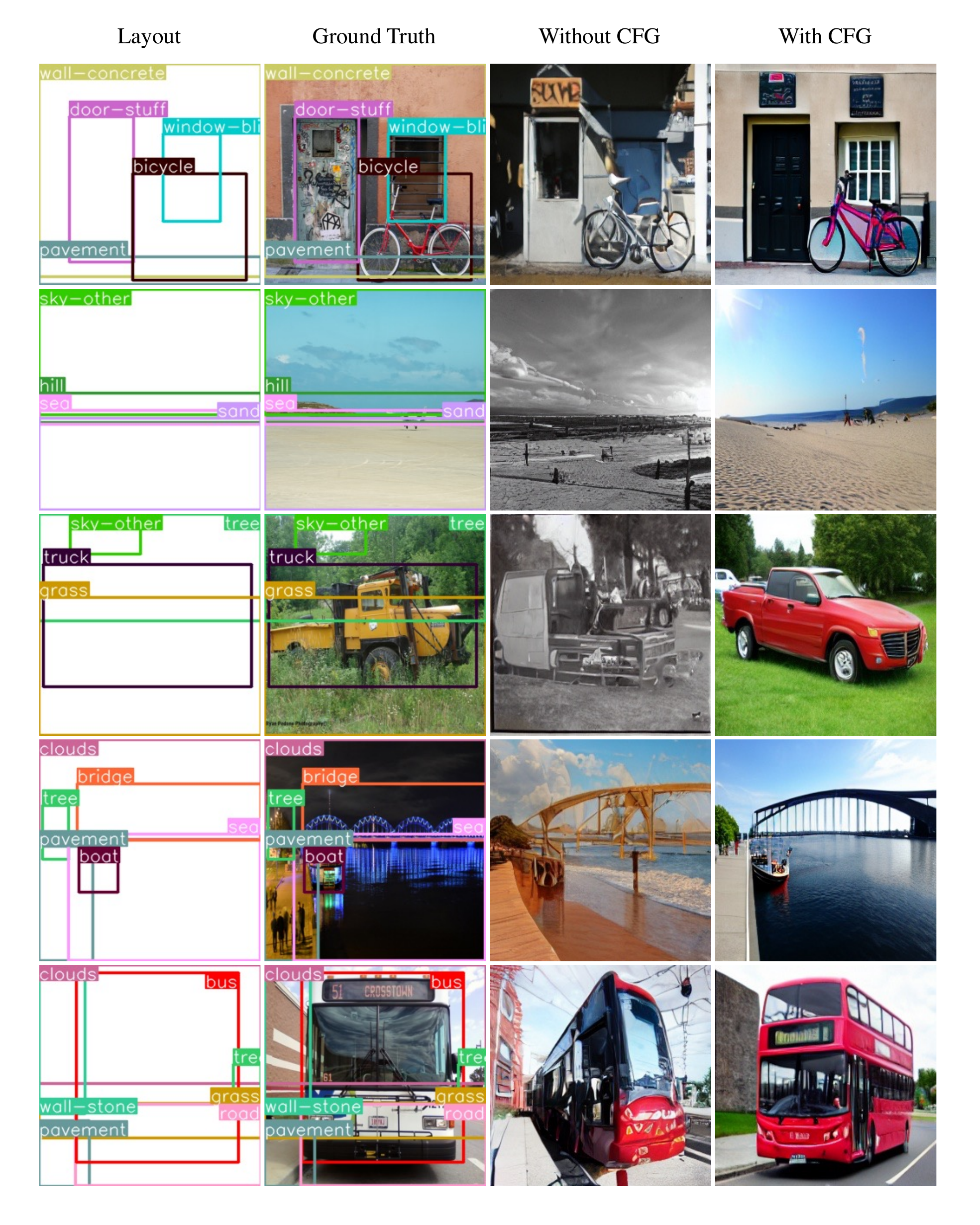}
    \caption{Classifier free guidance is important for high-quality image generation. By using ``null attention'', we are able to do CFG without a negative text prompt.}
    \label{fig.cfg}
\end{figure}

Classifier free guidance (CFG)~\cite{ho2022classifier} has shown its ability to improve the quality of images generated by conditional diffusion models (DMs). Typically, when applying CFG for a text-to-image DM, the model takes an empty text string as the input to form the negative condition. However, many applications of DM cannot take the advantage of CFG due to the absence of negative condition, \eg, unconditional DM~\cite{ddpm,beat_gan} and super-resolution DM~\cite{palette}. \ours\ can either be conditioned on text or be text free. When the model is text conditioned, we can freely use the empty string as the negative condition as a text-to-image DM. However, when using the text-free version of \ours, the generated image quality can be low without CFG. To solve this problem, we propose to feed empty layout and null embeddings (we call it ``null attention") to the model, which is described in~\cref{sec.instance_prompt_attn}, as the negative condition. \cref{fig.cfg} compares the bounding box layout-to-image generation with and without null-attention-based CFG on a text-free \ours. It is easy to see that the objects in images are barely recognizable without CFG and adding CFG greatly improves the image quality, object recognizability and layout fidelity.

\begin{figure}[!h]
    \centering
    \includegraphics[width=\linewidth]{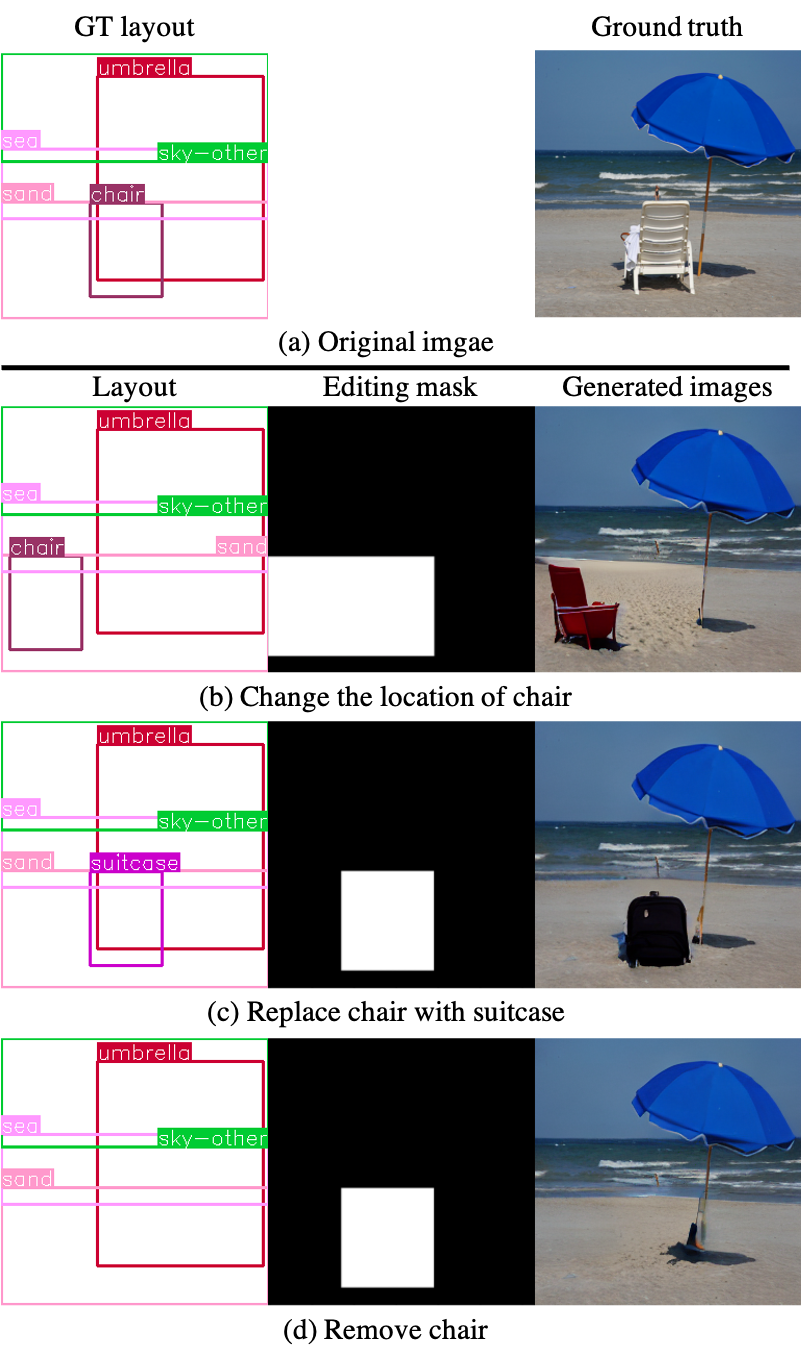}
    \caption{Image editing with \ours. (a) Ground truth layout and image. (b),(c) and (d) Edited layouts, editing regions (white) and generated images.}
    \label{fig.image_edit}
\end{figure}

\section{Image Editing with \ours}

\ours\ allows user to edit image by modifying the layout. We use a mask-driven method to edit image, which is similar to~\cite{palette,ldm}. During image generation, the model takes as input a binary mask (see \cref{fig.image_edit}). The mask indicates regions need modification and regions to remain. After each denoising step, the editing region is the output of DM while the remaining region is replaced by the diffused original image. 

We show three layout-guided image editing cases in \cref{fig.image_edit}. Our method can i) change the location of an object, ii) replace object with another object of different category, and iii) remove object in a region. In the first case, the editing mask is the union region of object before and after moving. Please note that \ours\ cannot maintain the identity of the object. However, it is possible to combine recent textual inversion fine-tuning methods~\cite{textual_inversion,dreambooth} to achieve identity preserving editing. We leave this to future work. In the latter two cases, the editing region is indicated by the bounding box of object to replace or remove.

\section{Qualitative Results of Data Efficiency Training}

\begin{figure*}
    \centering
    \includegraphics[width=\linewidth]{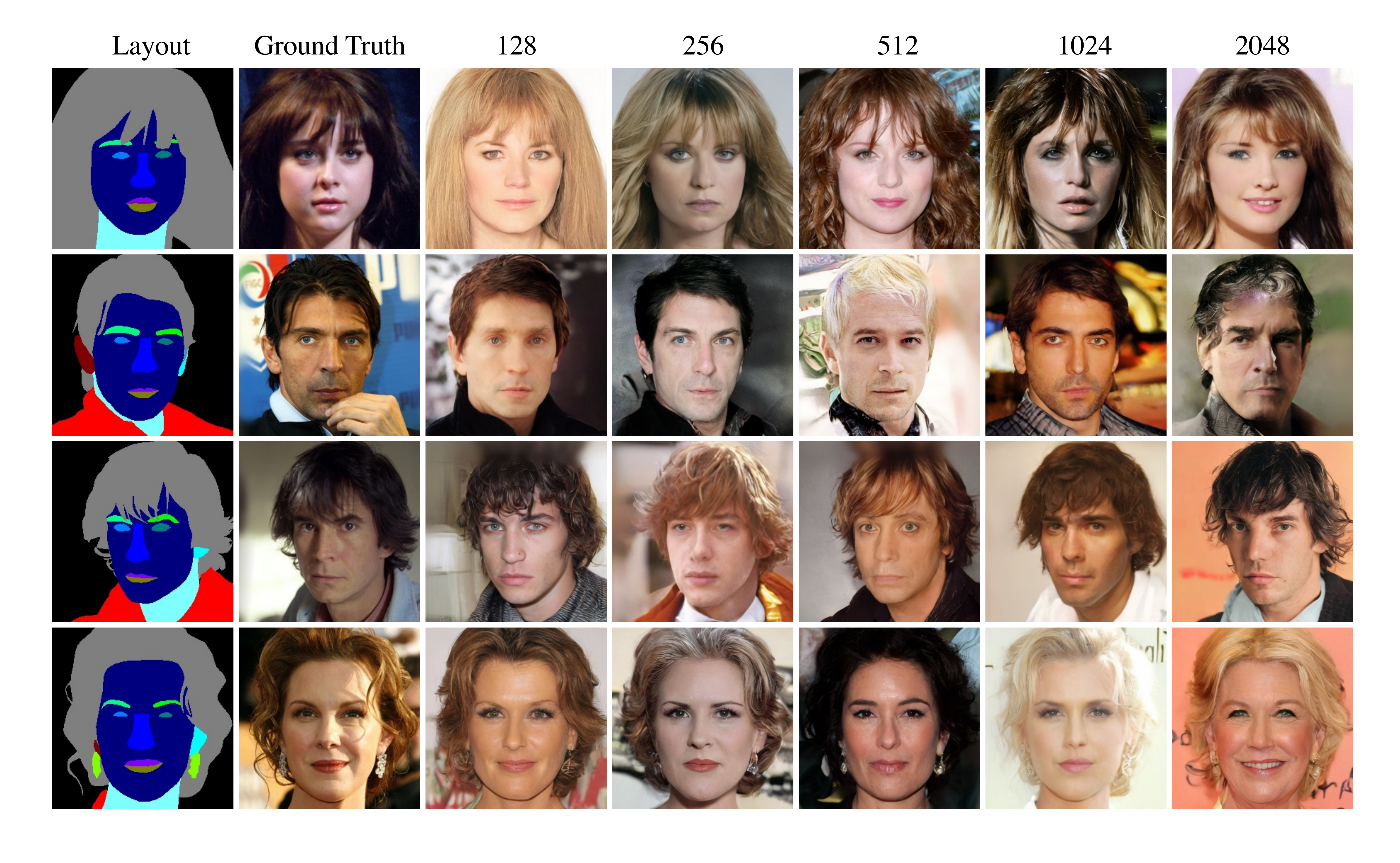}
    \caption{Qualitative results of models fine-tuned with 128, 256, 512, 1,024 and 2,048 images after 2,000 iterations. Despite trained with very few samples, the generated images align to layout well. The model also generates faces of different identities, indicating the model is not overfitted or collapsed.}
    \label{fig.data_effciency}
\end{figure*}

We quantitatively discussed the data efficiency in ~\cref{sec.efficiency}. \cref{fig.data_effciency} compares generation results of models that trained with 128, 256, 512, 1,024 and 2,048 samples. We can see that even with very few training sample, \ours\ is able to generate high-quality, diversified faces that aligns well to given layout.

\section{Limitation}
A limitation of our method is that the adapted models always have a larger size denoising model as we inject layout attention layers. \cref{tab.model_size} compares the denosing U-Net model size of foundational DM before fine-tuning and \ours. We notice that the number of U-Net model parameters only increases 1.69\% and 5.47\% when conditioned on bounding box and segmentation mask respectively, which will not cause computational bottleneck due to model size.

\begin{table}[!h]
    \centering
    \begin{tabular}{cccc}
    \toprule
    \multirow{2}{*}{Condition}& \multicolumn{3}{c}{Num. Params} \\
    \cmidrule(lr){2-4} 
     & Found. Model & \oursshort & Increase \\
    \cmidrule(lr){1-1} \cmidrule(lr){2-2}  \cmidrule(lr){3-3} \cmidrule(lr){4-4} 
    Bbox & 888M & 903M & 1.69\% \\
    Mask & 274M & 289M & 5.47\% \\
    \bottomrule
    \end{tabular}
    \caption{Model size comparison between the denoising U-Net of the foundational models and \ours.}
    \label{tab.model_size}
\end{table}

\begin{figure}
    \centering
    \includegraphics[width=0.98\linewidth]{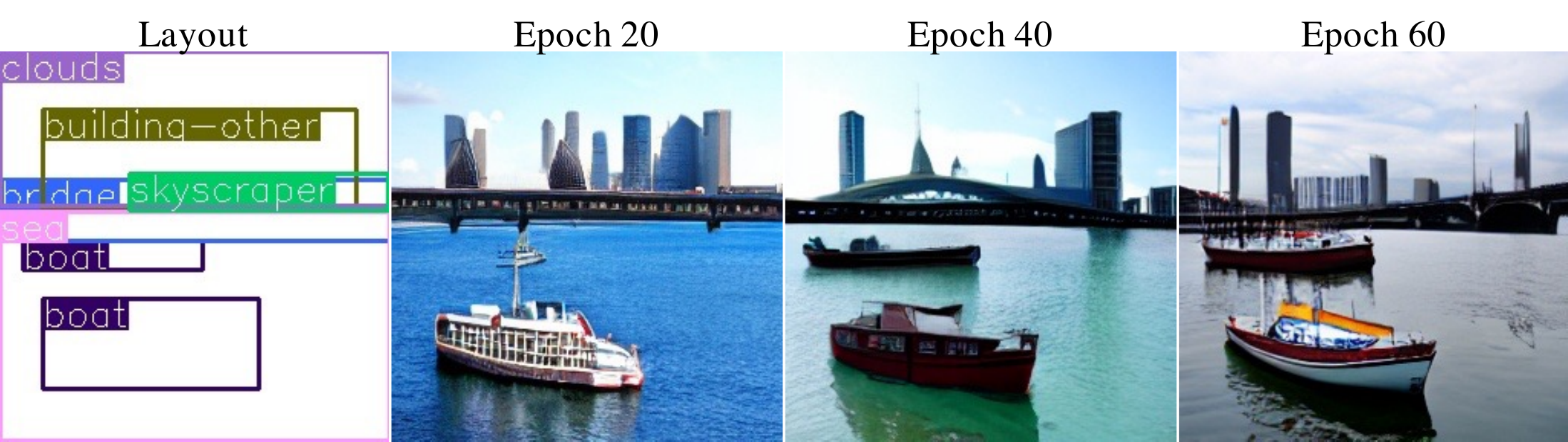}
    \caption{Qualitative visuals associated with \ref{fig.model_ablation}. Alignment accuracy increases with more epochs (see boats and clouds).}
    \label{fig:qualitative_fig_6}
\end{figure}

\begin{figure}
    \centering
    \includegraphics[width=0.98\linewidth]{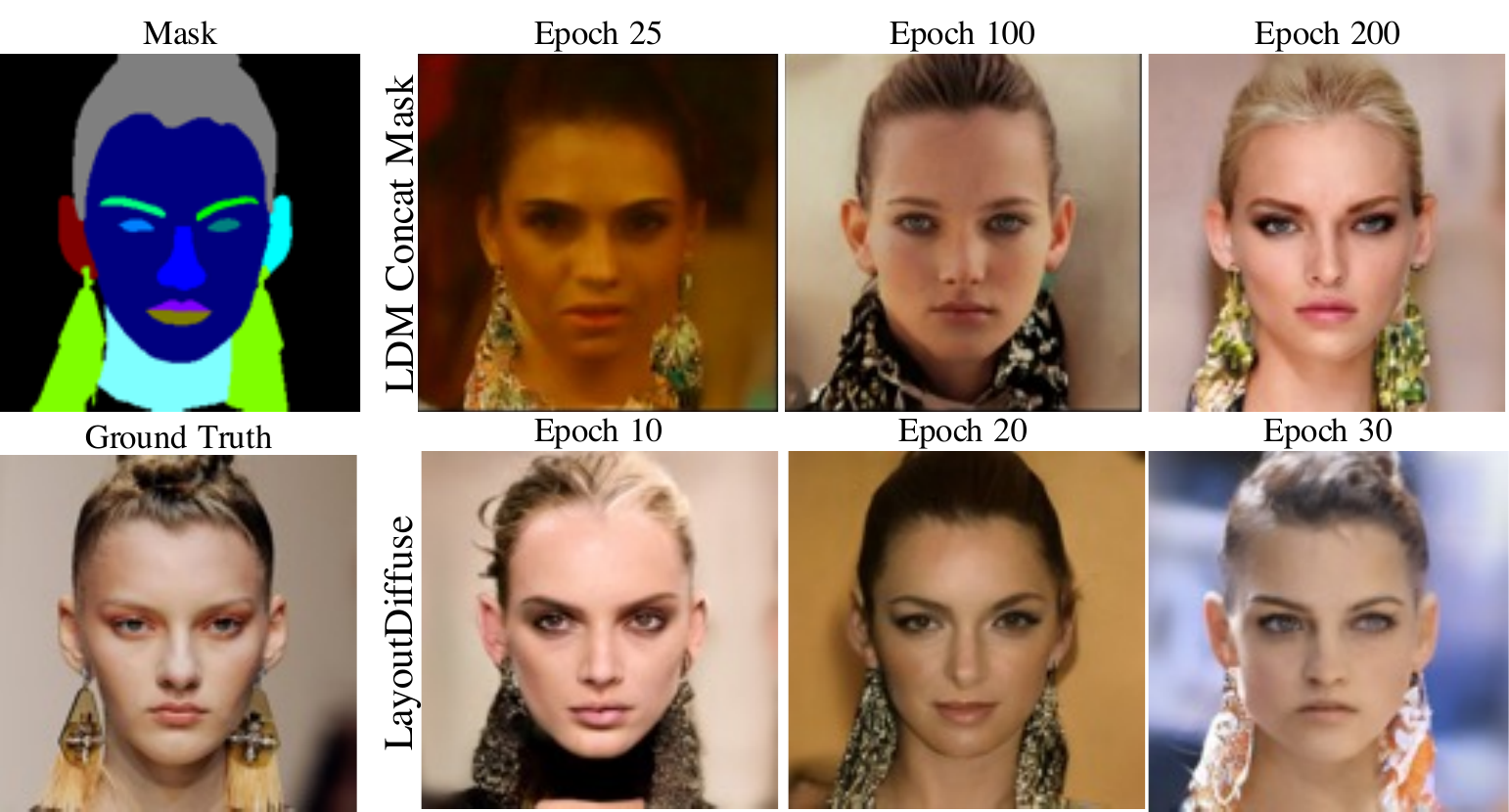}
    \caption{Qualitative visuals associated with \ref{fig.time_efficiency}. Concatenating mask to input takes much longer fine-tuning to achieve good results.}
    \label{fig:qualitative_figure_7}
\end{figure}

\section{More Qualitative Results}

\begin{figure}[!h]
    \centering
    \includegraphics[width=\linewidth]{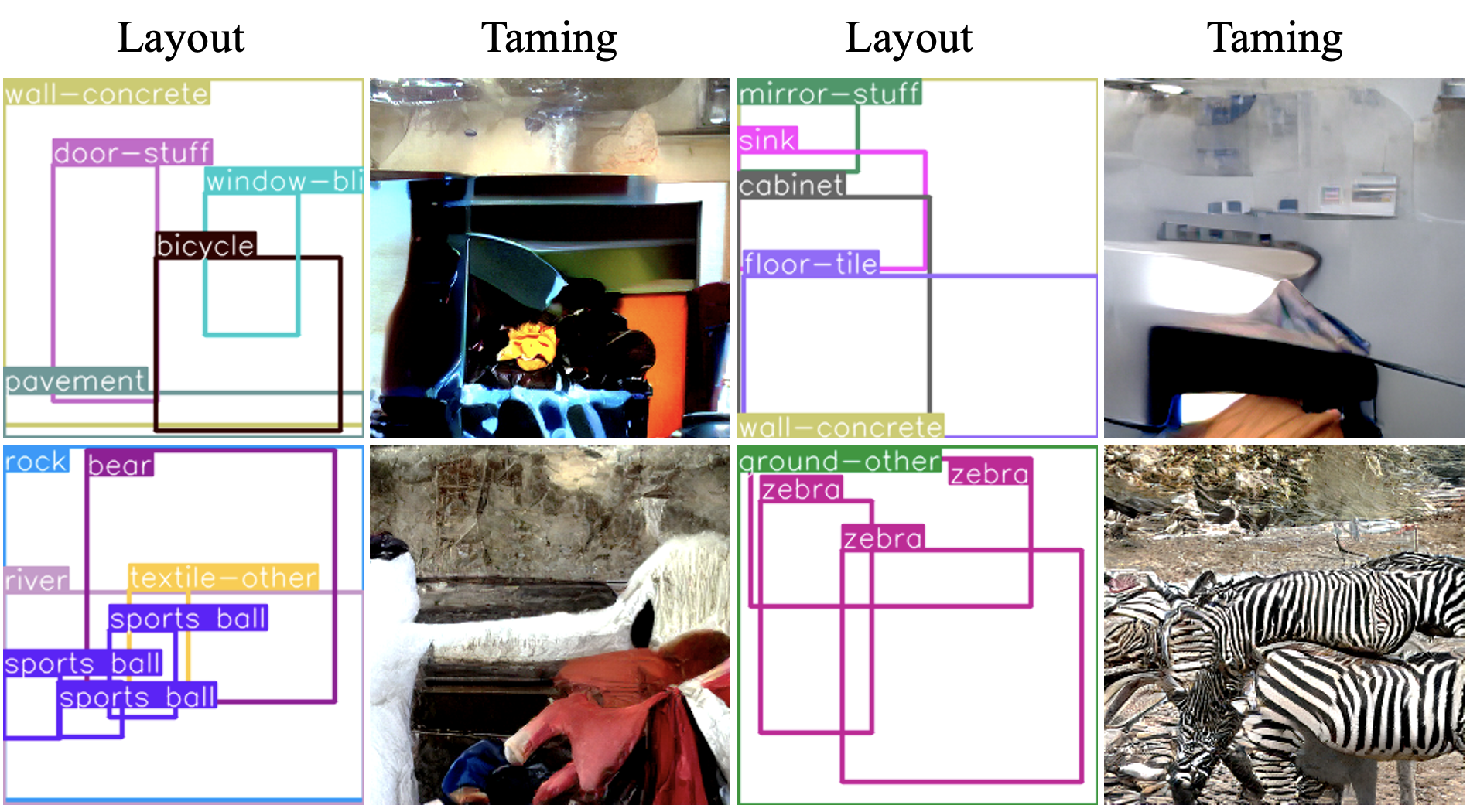}
    \caption{The third party code (including pre-trained model weights) for Taming baseline~\cite{taming} does not generate reasonable results. }
    \label{fig.taming}
\end{figure}

\begin{figure*}
    \centering
    \includegraphics[width=\linewidth,trim=0.3cm 0 0.4cm 0,clip]{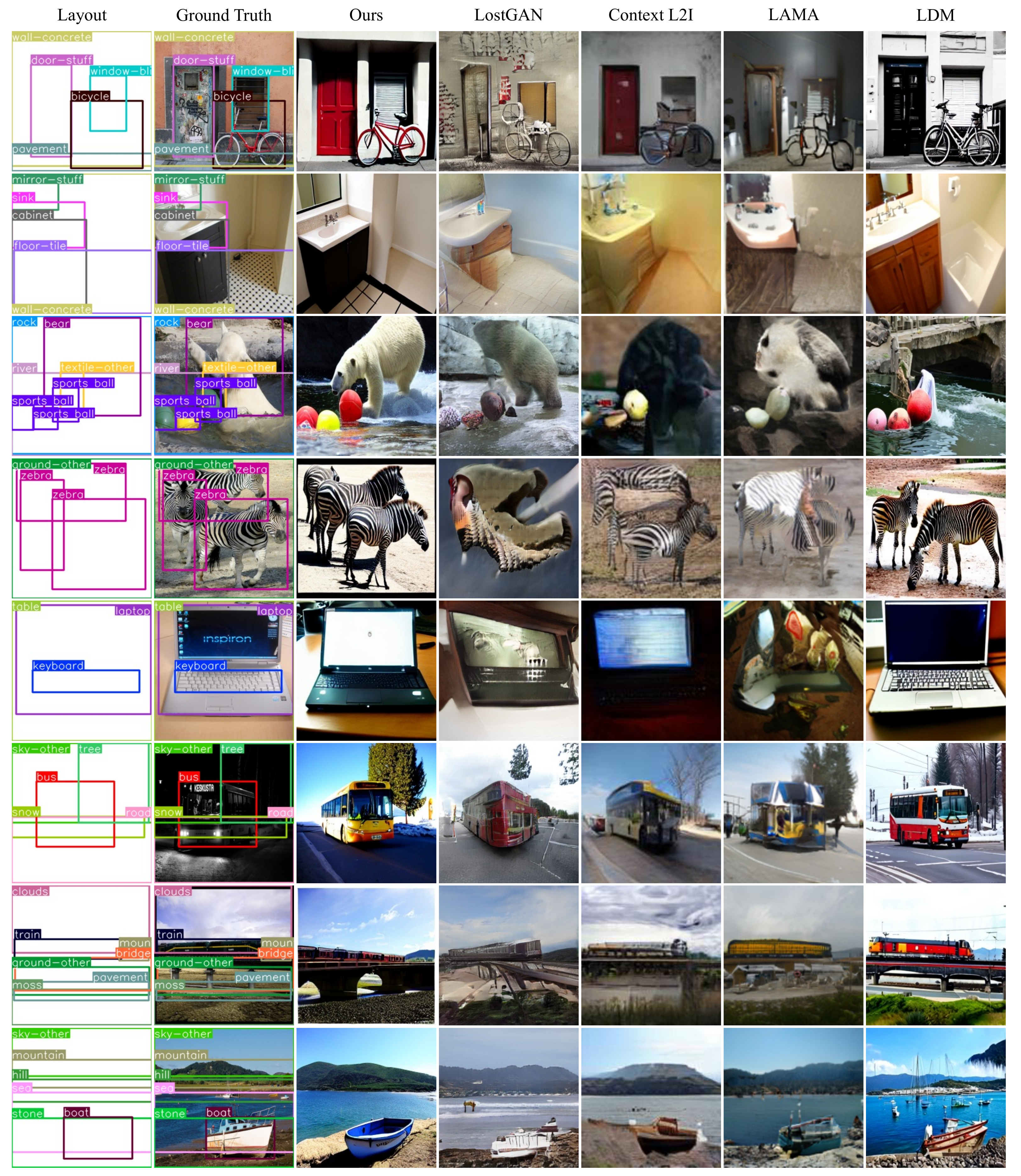}
    \caption{More qualitative result of bounding box layout-to-image generation on COCO dataset. Zoom in for better view.}
    \label{fig.appendix_qualitative}
\end{figure*}

\begin{figure*}
    \centering
    \includegraphics[width=\linewidth,trim=0.3cm 0 0.4cm 0,clip]{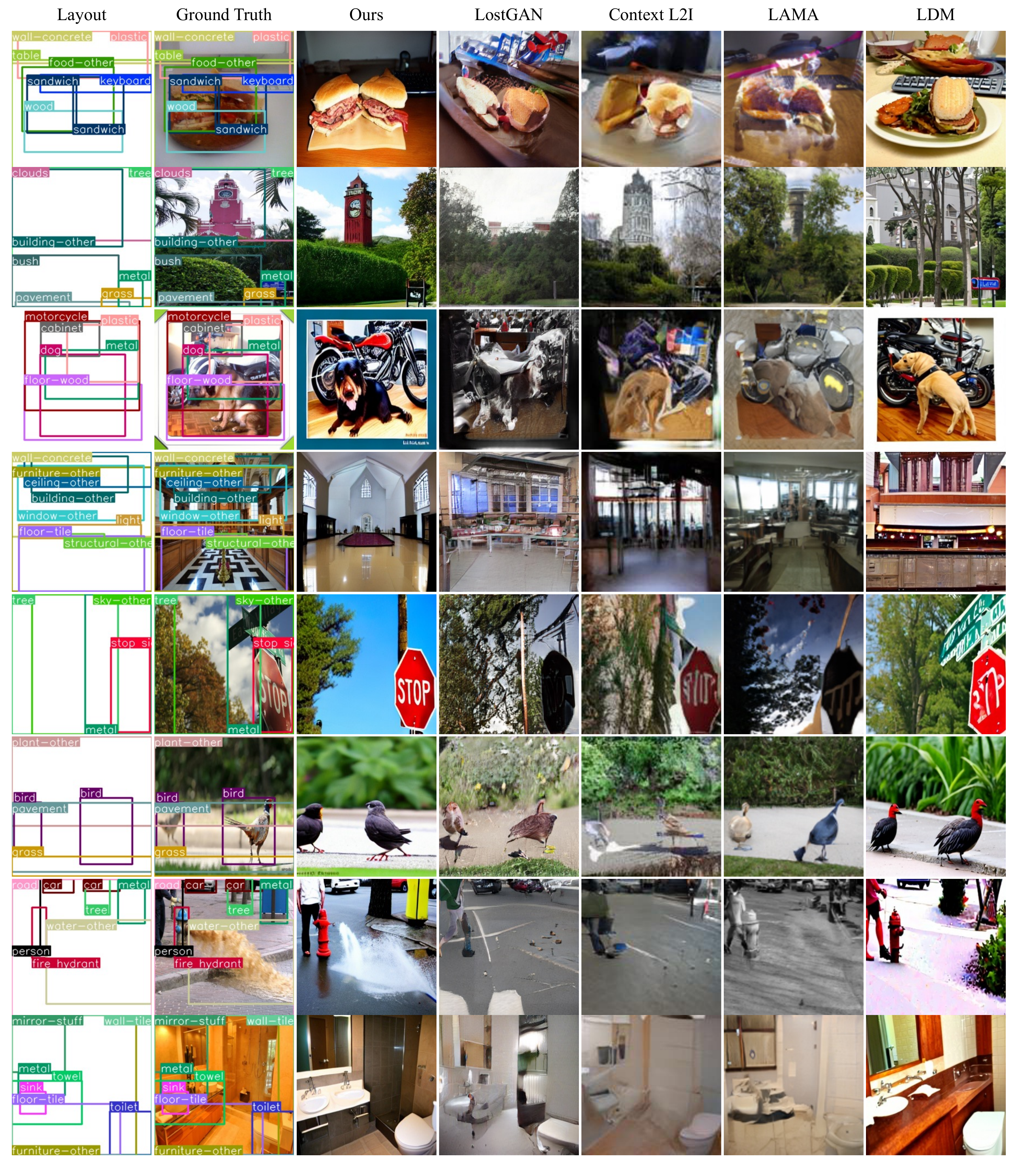}
    \caption{More qualitative result of bounding box layout-to-image generation on COCO dataset. Zoom in for better view.}
    \label{fig.appendix_qualitative2}
\end{figure*}

We provide more qualitative results in~\cref{fig.appendix_qualitative,fig.appendix_qualitative2}. Our results have not included figures from VQ-VAE+AR methods because i) we are not able to obtain runnable code from TwFA~\cite{twfa}, the code releasing  repository\footnote{https://github.com/JohnDreamer/TwFA} is empty when we submit this work, and ii) the only available implementation\footnote{https://github.com/CompVis/taming-transformers\#scene-image-synthesis} for Taming~\cite{taming} is from third party and does not generate reasonable figures (see \cref{fig.taming}).

\end{document}